\documentclass[twoside,leqno,twocolumn]{article}

\usepackage[letterpaper]{geometry}

\usepackage{ltexpprt}
\usepackage{hyperref}
\usepackage[dvipsnames]{xcolor}
\usepackage{booktabs}
\usepackage{graphicx}
\usepackage{float}
\usepackage{multirow}

\begin{document}

\newcommand\relatedversion{}



\title{\Large Beyond Fairness: Age-Harmless Parkinson's Detection via Voice}

\author{Yicheng Wang\thanks{Department of Computer Science and Engineering, Texas A\&M University, \{wangyc, han\}@tamu.edu.}
\and Xiaotian Han$^{*}$
\and Leisheng Yu\thanks{Department of Computer Science, Rice University, leisheng.yu@rice.edu.}
\and Na Zou\thanks{Department of Engineering Technology and Industrial Distribution, Texas A\&M University, nzou1@tamu.edu.}
}

\date{}

\maketitle


\fancyfoot[R]{\scriptsize{Under Review}}





\begin{abstract} \small\baselineskip=9pt

Parkinson's disease (PD), a neurodegenerative disorder, often manifests as speech and voice dysfunction. While utilizing voice data for PD detection has great potential in clinical applications, the widely used deep learning models currently have fairness issues regarding different ages of onset. These deep models perform well for the elderly group (age $>$ 55) but are less accurate for the young group (age $\leq$ 55). Through our investigation,  the discrepancy between the elderly and the young arises due to 1) an imbalanced dataset and 2) the milder symptoms often seen in early-onset patients. However, traditional debiasing methods are impractical as they typically impair the prediction accuracy for the majority group while minimizing the discrepancy. To address this issue, we present a new debiasing method using GradCAM-based feature masking combined with ensemble models, ensuring that neither fairness nor accuracy is compromised. Specifically, the GradCAM-based feature masking selectively obscures age-related features in the input voice data while preserving essential information for PD detection. The ensemble models further improve the prediction accuracy for the minority (young group). Our approach effectively improves detection accuracy for early-onset patients without sacrificing performance for the elderly group. Additionally, we propose a two-step detection strategy for the young group, offering a practical risk assessment for potential early-onset PD patients.

\end{abstract}
\vspace{5pt}

\section{Introduction}

Parkinson's Disease (PD) is a progressive neurodegenerative disorder, with a significantly increasing mortality over the decades \cite{rong2021trends}. Common symptoms of PD include speech changes, termed as hypokinetic dysarthria. These changes present as hypophonia (reduced voice volume), dysphonia (impaired voice quality), and hypoprosodia (diminished pitch inflection) \cite{sapir2008speech}, etc. Previous studies have showed that over $89\%$ \cite{ho1998speech, schalling2018speech, ramig2018speech} PD patients develop speech disorders. Accordingly, voice-based PD detection methods offer promising prospects for application in clinical PD diagnosis \cite{ngo2022computerized}. Doctors can perform preliminary screenings swiftly, conveniently, and even remotely for individuals suspected of having PD, thus facilitating early detection and treatment. Several software systems \cite{ozkan2016comparison, zhang2020intelligent} have been developed specifically for PD tele-diagnosis. 

\begin{table}[t]
    \centering
    \setlength{\tabcolsep}{11pt}
    \caption{The AUPRC performance of the models \cite{karaman2021robust} on the mPower Dataset \cite{bot2016mpower}: Young Group vs. Elderly Group. {\color{red}$\Delta$} shows the AUPRC difference between the young and elderly groups.}
    \label{table1}
    \begin{tabular}{l c c c}
        \toprule
        Model       &  Young & Elderly & {\color{red}$\Delta$} \\
        \midrule
        ResNet50       &  0.6517 & 0.9739 & {\color{red}0.3222}  \\
        DenseNet161    &  0.6735 & 0.9736 & {\color{red}0.3001}  \\
        \bottomrule
    \end{tabular}

\end{table}

PD voice detection methods based on machine learning techniques, such as logistic regression (LR), Support Vector Machines (SVMs), and Random Forests (RF) \cite{tai2021voice, mohammadi2021parkinson, polat2020parkinson}, achieve fast and accurate results. Recently, methods based on deep learning such as Convolutional Neural Networks (CNNs)\cite{er2021parkinson, karaman2021robust} have dominated PD voice detection with their efficacy. However, as demonstrated in previous studies \cite{mehrabi2021survey,caton2020fairness,nanda2021fairness}, deep learning methods can exhibit biases towards certain groups. Despite this, only a few of studies \cite{rahman2021detecting} have considered the fairness issue, particularly concerning early-onset PD (EOPD) and late-onset PD (LOPD). To demonstrate the bias inherent in deep learning PD detection methods, we conducted preliminary experiments using the state-of-the-art CNN models \cite{karaman2021robust}. Specifically, we investigated the performance disparities between the young and elderly groups. The results in Table \ref{table1} display the biases towards detecting EOPDs and LOPDs on the mPower PD dataset \cite{bot2016mpower}. While these models exhibit robust performance for the elderly group (age $> 55$), they are less effective for the young group (age $\leq 55$) with a significant disparity in the AUPRC (Area Under the Precision-Recall Curve) metric. 

From our preliminary investigation, we identified two possible reasons behind the performance discrepancy. Firstly, the issue may arise from imbalanced data distribution. Given that PD is more prevalent among the elderly, the proportion of PD patients in the young group is significantly lower than in the elderly one. This can make PD detection models more likely to overfit when assessing young individuals. The second reason is that the voice characteristics of EOPDs are not as obvious as those of LOPDs. There are evident phenotype-specific speech distinctions between the two groups, as highlighted by Rusz et al. \cite{rusz2021distinct}. We have also observed that many EOPD cases show less pronounced features than LOPDs. 

To mitigate the bias in PD prediction, various general-purpose bias mitigation methods are available, including resampling \cite{chakraborty2021bias, idrissi2022simple}, reweighting \cite{krasanakis2018adaptive}, and adversarial learning \cite{zhang2018mitigating}. However, such approaches cannot solve the overfitting problem or typically sacrifice the models' performance on the primary group to bridge the performance gap between groups, in other words, result in a performance decline for the elderly group. Given our task of PD detection, any compromise in performance for the majority (elderly) group is unacceptable. Therefore, achieving fair voice-based detection for both early and late-onset PDs without harming performance for the elderly group continues to be a challenge.

To address this, we introduce a novel debiasing method that incorporates GradCAM-based feature masking along with ensemble modeling, applied specifically to CNN models designed for PD voice detection. With the aim to remove age-related information in the input features,  we propose to mask age-related features and harness the capabilities of multi-models. Our proposed approach effectively narrows the performance (AUPRC) gap between the young and elderly groups without compromising the performance of the elderly group. We also develop a two-step detection strategy for the young group. This supplementary detection phase serves as a screening method, yielding more precise detection results and a risk assessment for subsequent PD diagnosis.






We highlight our main contributions as follows:

\begin{itemize}
    \item \textbf{Sources Identification of Model Bias:} We identify the key factors responsible for the differential model performance observed between EOPDs and LOPDs, notably the challenges arising from imbalanced data distribution and the nuanced characteristics inherent to EOPDs.
    \item \textbf{Innovative Debiasing Method for PD Voice Detection:} Based on the above finding, we propose a novel debiasing approach employing GradCAM-based masking and ensemble models to address the performance discrepancy between the young and elderly groups. We also illustrate that other debiasing methods cannot resolve the overfitting issue or they unavoidably sacrifice overall performance.
\item \textbf{Two-Step PD Voice Detection Strategy:} We introduce a refined two-step strategy for EOPD prediction and risk assessment, which is more effective and practical in real-world clinical applications.
\end{itemize}

\section{Related Work}
This section introduces the two main lines of work related to our study, including machine learning methods for PD voice prediction and fairness in healthcare. 

\paragraph{Machine Learning for PD Voice Prediction.} While the use of a patient's voice and speech signal for detecting and screening PD has not yet become a standard clinical diagnostic tool, it holds considerable promise for future applications \cite{ngo2022computerized}. A range of machine learning \cite{tai2021voice, mohammadi2021parkinson, polat2020parkinson} and deep learning models \cite{er2021parkinson, karaman2021robust} have been employed for this purpose. One study \cite{karaman2021robust} utilized pretrained CNN models, including SqueezeNet, ResNet, and DenseNet, with time–frequency domain mel-spectrogram as input data. This approach achieved approximately $0.9$ in accuracy, recall, and precision scores. Furthermore, voice signals can be complemented with other actionable data, such as facial features, for PD detection \cite{lim2022integrated}. Several software solutions \cite{ozkan2016comparison, zhang2020intelligent} have been developed for remote PD diagnosis. In parallel, researchers have introduced numerous PD voice datasets \cite{orozco2014new, bot2016mpower, sakar2019comparative} to facilitate the development of new methods and data analysis. The comprehensive mPower database \cite{bot2016mpower} stands out as the largest public PD voice dataset, suitable for training deep neural networks.

\paragraph{Fairness in Healthcare.}
The healthcare domain is high-stakes for machine learning methods~\cite{seastedt2022global,ahmad2020fairness,chen2021algorithm}, as they can exhibit biases towards certain groups. Such biases are unacceptable in healthcare for fairness reasons. Previous studies have extensively explored and highlighted this issue. Ding et al. \cite{ding2022fairly} proposed a fair machine learning framework targeting graft failure prediction in liver transplants, and Chang et al. \cite{chang2023towards} proposed a fair patient-trial matching framework by generating a patient-criterion level fairness constraint. Moreover, while enforcing fairness in machine learning for healthcare is critical, it can also be detrimental to healthcare tasks. It is essential to aim for models that achieve as much fairness as possible without causing undue harm to any subgroup \cite{ustun2019fairness}. Martinez et al. \cite{martinez2019fairness} suggested achieving 'Fairness With Minimal Harm' using a Pareto-Optimal approach, which dynamically rebalances the risks of subgroups.

\begin{table}[t]
    \centering
    \setlength{\tabcolsep}{9pt}
    \caption{PD/HC label distribution in the mPower dataset: Young Group vs. Elderly Group. The PD/HC ratio in the young group is much more imbalanced than in the elderly group.}
    \begin{tabular}{l c c c}
        \toprule
        Group &  \#PD & \#HC & \#PD/\#HC\\
        \midrule
        Young ($\leq55$)  &  $1,394$ & $17,528$ & 0.0795\\
        Elderly ($>55$)   &  $9,175$ & $5,373$ & 1.7076\\
        \bottomrule
    \end{tabular}
    \label{table2}
\end{table}

\section{Preliminaries}
In this paper, we aim to address fairness issues associated with the mPower dataset \cite{bot2016mpower} in the state-of-the-art method for PD screening, i.e., CNN-based approach (ResNet50 and DenseNet161)\cite{karaman2021robust}. We follow the data preparation in Karaman et al.'s study \cite{karaman2021robust} using the $10$-second "a$\sim$" mel-spectrogram data from the mPower dataset \cite{bot2016mpower}. The goal is to categorize individuals as either Parkinson's Disease (PD) patients or Healthy Controls (HC). We allocate the dataset into a $4:1$ ratio for the training and testing sets, totally consisting of $10,569$ PD cases and $22,901$ HC cases. While a universally agreed age boundary definition for EOPD and LOPD is yet to be confirmed, previous studies typically placed it between 40 and 65 years of age \cite{quinn1987young, schrag2006epidemiological, giovannini1991early, camerucci2021early, gibb1988comparison, mehanna2019young}. Based on these, we set a cut-off age of 55 to distinguish the young group and the elderly group \cite{camerucci2021early}.

\begin{figure}[!t]
    \centering
    \includegraphics[width=\columnwidth]{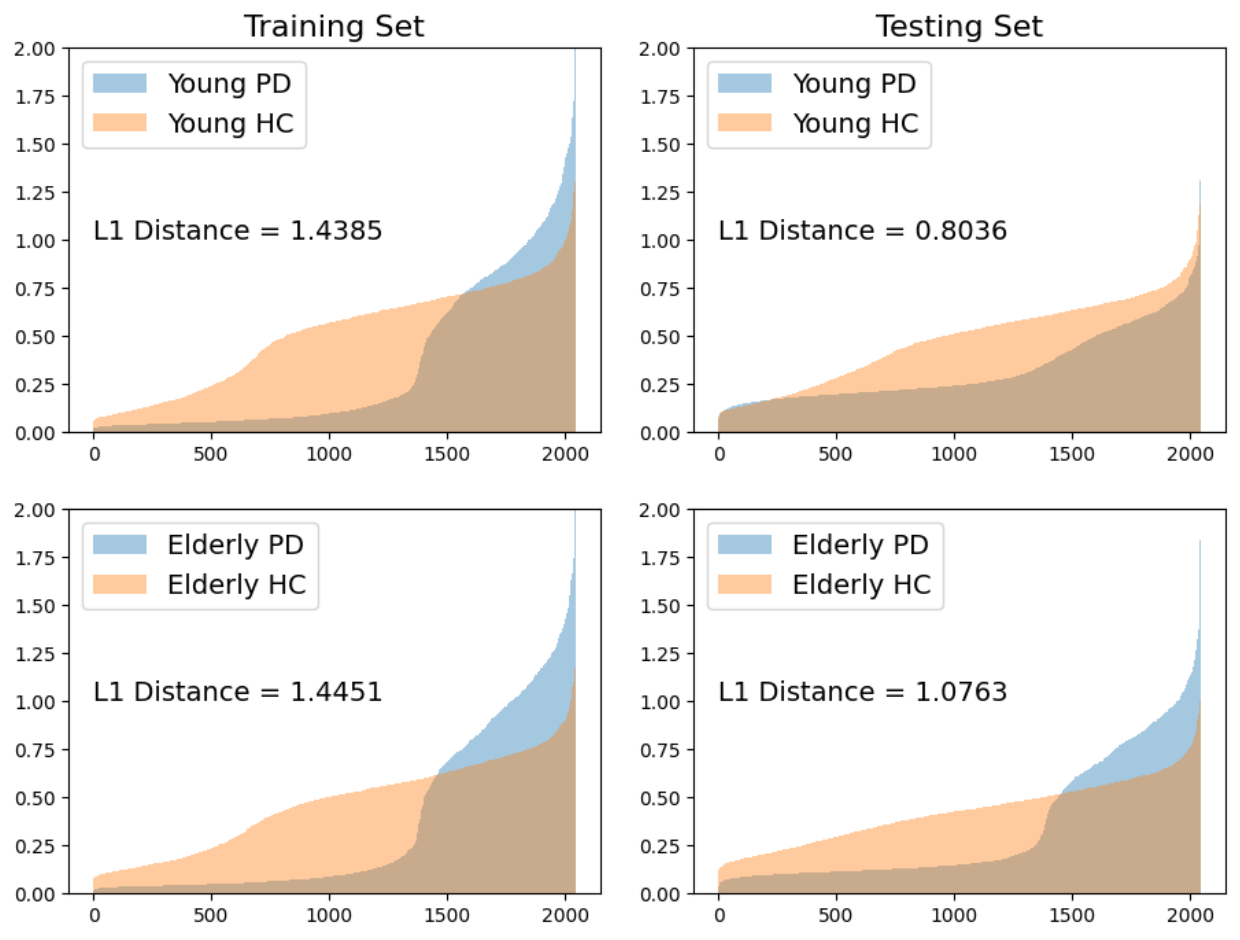}
    \vspace{-15pt}
    \caption{A comparison of averaged and sorted ResNet50 features between the training and testing sets. Consistent and notably large L1 distances for both age groups in the training set suggest the model's proficiency but this also reveals overfitting while considering the testing set, especially pronounced in the young group with a smaller L1 distance.}
    \label{fig1}
\end{figure}

\subsection{Imbalanced Data Distribution.}

We first investigate the data distribution among the young and elderly groups. Not surprisingly, because PD usually occurs in the elderly population and is relatively rare in young individuals, the PD/HC ratio is extremely imbalanced in the young group as shown in Table \ref{table2}. Previous studies on various tasks \cite{johnson2019survey, subramanian2021fairness, buda2018systematic, li2020analyzing} have shown that such class imbalance in datasets can lead to a significant decline in model performance and cause overfitting problems. 

We employ the averaged and sorted output features from different groups to illustrate overfitting in PD detection on the mPower dataset. Features from the final layer (2048-dimensional) are extracted from a well-trained ResNet50 for all data, then averaged and sorted. We compute the L1 distance between the PD and HC features for both the young and elderly groups. As depicted in Figure \ref{fig1}, the model performs well on the training set, indicated by the large L1 distances for both age groups. The features distinguishing PD from HC are quite pronounced for both young and elderly groups. However, on the testing set, the L1 distance between PD and HC for the young group is noticeably smaller than that for the elderly group and also significantly reduced compared to the training set. Clearly, the features of PD and HC in the young group from the testing set are more similar, leading to a higher likelihood of misclassifying EOPD cases. In conclusion, the model exhibits overfitting on the dataset, particularly for the young group, due to the imbalanced data distribution.

\begin{figure}[!t]
    \includegraphics[width=\columnwidth]{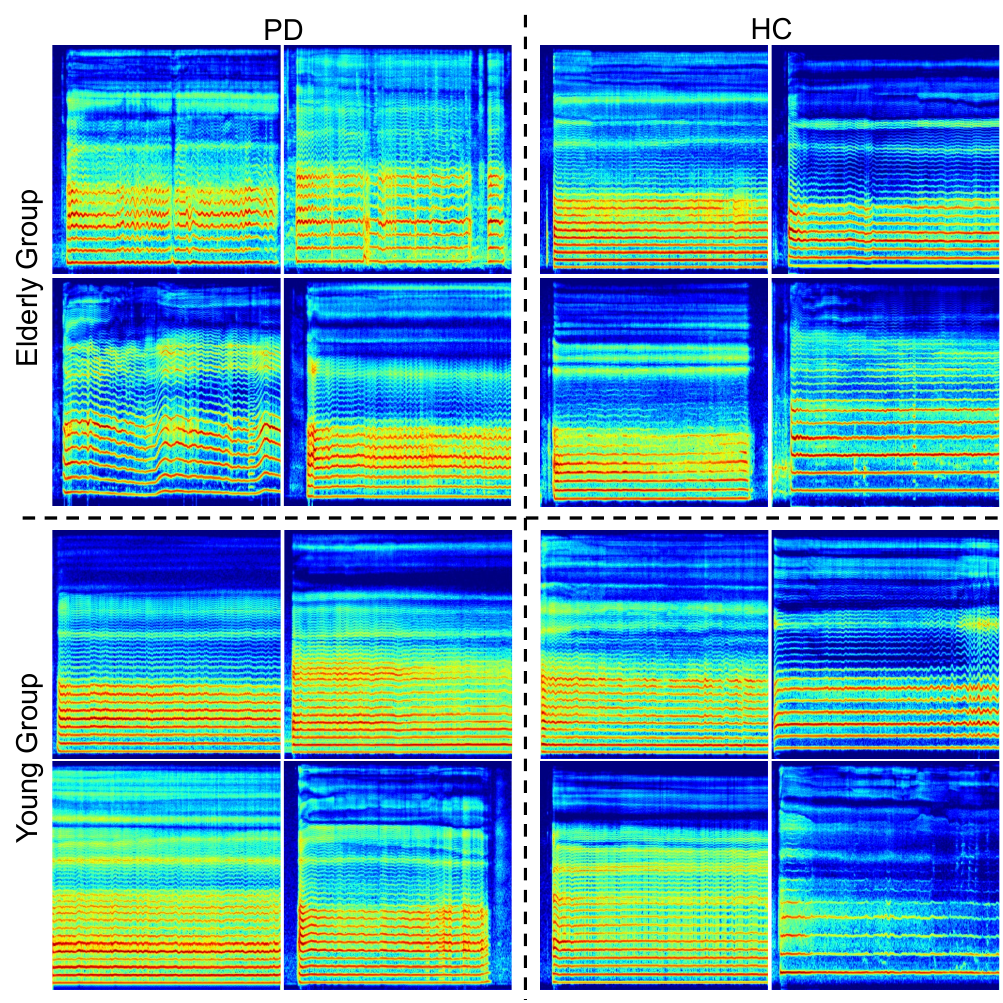}
    \vspace{-15pt}
    \caption{Exemplar PD and HC mel-spectrograms in the elderly and young groups. It can be observed that, compared to HCs, the features of PDs in the elderly group are more pronounced than those in the young group.}
    \label{fig2}
\end{figure}

\begin{figure*}[!t]
    \centering
    \includegraphics[width=0.9\textwidth]{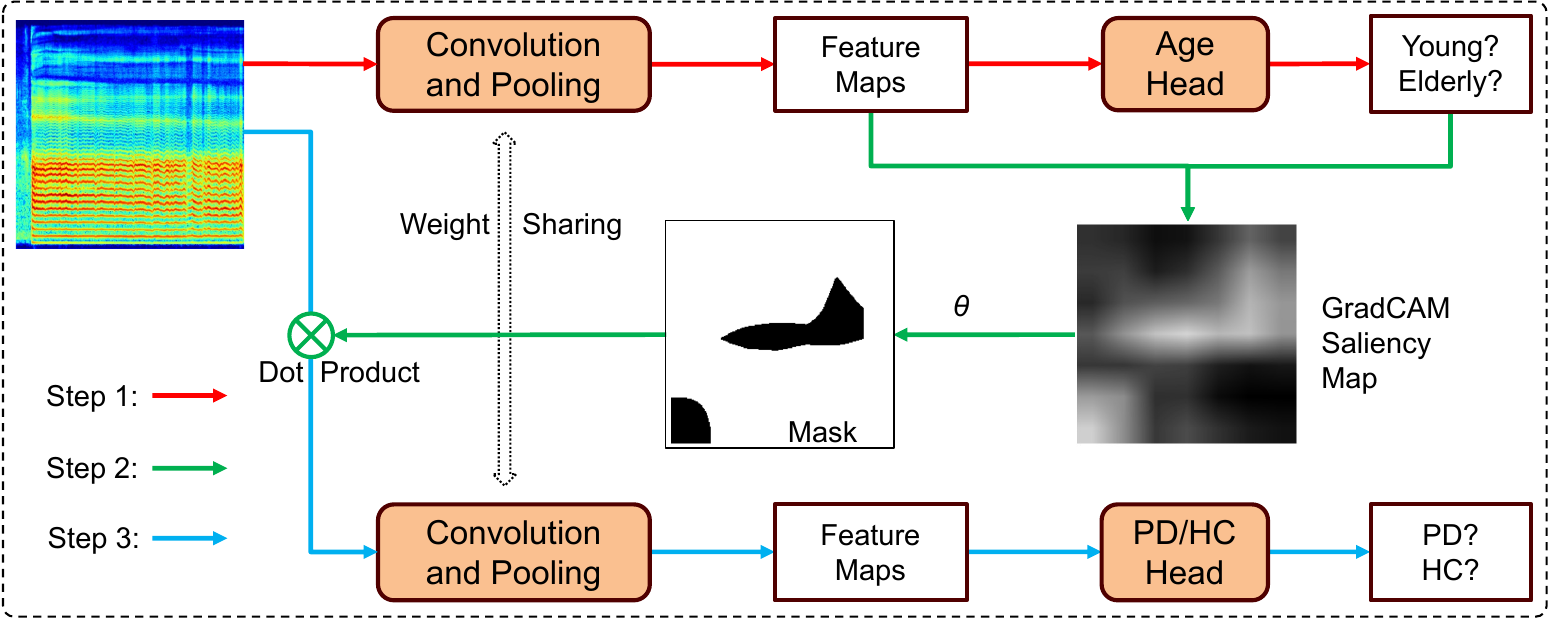}
    \caption{Overview of the proposed GradCAM-based feature masking pipeline. \textbf{{\color{red} Step 1 ($\longrightarrow$)}}: Utilize a CNN model to extract features from the input mel-spectrograms, followed by predicting the age group using a binary classification head (age head).
    \textbf{{\color{ForestGreen} Step 1 ($\longrightarrow$)}}: Produce the GradCAM saliency map \cite{selvaraju2017grad} from the age prediction and feature maps. Subsequently, apply a threshold to generate the GradCAM-based mask and overlay this mask onto the original input. 
    \textbf{{\color{Cyan} Step 3 ($\longrightarrow$)}}: From the masked input, extract features with the same CNN model and then predict the PD/HC designation using another binary classification head (PD/HC head).}
    \label{fig3}
\end{figure*}

\subsection{Case Study on PD/HC Data.}

To gain more insights into the PD prediction task, we conducted a case study on PD/HC data.  We display several cases of PD and HC from both the elderly and young groups in Figure \ref{fig2}. Each image is a mel-spectrogram of a single datapoint, serving as the initial input of the CNN-based PD detection models. The x-axis of each mel-spectrogram corresponds to time (seconds 0 to 10), while the y-axis represents frequency in the mel scale \cite{stevens1937scale, o1987speech}. Each image is rendered from a $0-1$ time-frequency 2D array employing jet color maps with a size of $3\times224\times224$. Within the elderly group, the PD data reveal more pronounced tremors, jitters, and interruptions, aligning with typical voice and speech symptoms observed in PD patients. In contrast, young PD cases shown in the figure closely resemble those of young HCs, suggesting that EOPDs may lack distinctive features like LOPDs. This similarity may arise due to the slower progression and milder symptoms of EOPDs \cite{ferguson2016early}, leading to challenges for the model in distinguishing EOPDs from healthy young individuals.

Our preliminary studies uncover two primary sources of bias in PD detection models: \textbf{data imbalance} and \textbf{understated features of EOPD}. Driven by these findings, our aim is to encourage the model to better capture the consistent characteristics present in both EOPD and LOPD cases.

\section{Methodology}
In this section, we present our method for fair PD voice detection,  GradCAM-based feature masking, and ensemble models. The basic ideas of the proposed method are 1) employing GradCAM-based feature masking to remove age-related features in the input and 2) utilizing ensemble models to leverage the capabilities of multiple models, thereby enhancing prediction accuracy and mitigating the issue of overfitting.

\begin{table*}[!t]
    \centering
    \setlength{\tabcolsep}{11pt}
    \caption{AUPRC results for the entire mPower dataset (Average), the young group (Young), the elderly group (Elderly), and the disparity between groups (Elderly-Young) across all methods. An upward arrow $\uparrow$ indicates that higher values are preferable, whereas a downward arrow $\downarrow$ signifies that lower values are desired. {\color{red}$\Delta$} shows the AUPRC difference between the young and elderly groups. The results show that our proposed method achieve the lowest {\color{red}$\Delta$}, indicating our method achieves fair but harmless performance.}
    \begin{tabular}{l c c c c c}
        \toprule
        Model &  Method & Average($\uparrow$) & Young($\uparrow$) & Elderly($\uparrow$) &  {\color{red}$\Delta$} ($\downarrow$)\\
        \midrule
        \multirow{6}{*}{ResNet50} &  Original Model & $0.9365$ & $0.6517$ & $0.9739$ &$0.3222$\\
        &  Resampling & $0.9366$ & $0.6710$ & $0.9714$ &$0.3004$\\
        &  Adversarial & $0.9059$ & $0.6728$ & $0.9596$ &$0.2868$\\
        &  GradCAM Mask& $0.9392$ & $0.6912$ & $0.9727$ &$0.2814$\\
        &  Ensemble & $0.9509$ & $0.6994$ & $0.9806$ &$0.2811$\\
        &  GradCAM Mask + Ensemble & $\mathbf{0.9564}$ & $\mathbf{0.7488}$ & $\mathbf{0.9810}$ &$\mathbf{0.2322}$\\
        
        \midrule

        \multirow{6}{*}{DenseNet161}  &  Original Model & $0.9388$ & $0.6735$ & $0.9736$ &$0.3001$\\
        &  Resample & $0.9408$ & $0.6675$ & $0.9752$ &$0.3077$\\
        &  Adversarial & $0.9100$ & $0.6764$ & $0.9606$ &$0.2842$\\
        &  GradCAM Mask& $0.9533$ & $0.6947$ & $0.9804$ &$0.2857$\\
        &  Ensemble & $0.9521$ & $0.7221$ & $0.9792$ &$0.2571$\\
        &  GradCAM Mask + Ensemble & $\mathbf{0.9663}$ & $\mathbf{0.7541}$ & $\mathbf{0.9869}$ &$\mathbf{0.2327}$\\
        \bottomrule
    \end{tabular}
    \label{table3}
\end{table*}

\subsection{GradCAM-based Feature Masking.}
Driven by the primary factors previously mentioned that lead to fairness concerns in PD voice detection, we develop a GradCAM-based feature masking pipeline. The proposed method aims to perform in-processing debiasing of the CNN model during its training phase via masking the age-related features. This ensures that the model is less likely to develop biases based on age. The detailed procedure is shown in Figure \ref{fig3}.

During training, we initially utilize a pre-trained CNN model, such as ResNet50, to extract feature maps from the input data. The feature maps are subsequently reshaped into vectors. We adopt a binary classification head (age head, a fully connected layer with an output size of 2) to predict the age groups of the input data. By computing the gradients of the age group predictions with respect to the feature maps, we derive the GradCAM saliency maps, $S$ \cite{selvaraju2017grad}.  These saliency maps are then converted into binary input feature masks, $M$, using a threshold value, $\theta$. Specifically, $M_{h,w}$ is set to 0 if $S_{h,w}$ exceeds $\theta$, and 1 otherwise. The original inputs are element-wise multiplied with the masks, effectively masking out age-related regions within the input data.  Finally, the masked inputs are fed back into the same CNN model, this time aiming to predict PD/HC using a different head. 

The loss function we adopt is the sum of two cross-entropy losses from both age and PD/HC predictions. The fundamental objective of this pipeline is to deter the model from capitalizing on features that also correlate with age prediction. Through the utilization of GradCAM saliency maps extracted from age prediction outcomes, our pipeline strategically eliminates age-centric features, directing the model's focus towards features that remain consistent across different age groups.

\subsection{Ensemble Models.}
To further enhance prediction accuracy for the young group, we utilize the capability of ensemble models. Ensemble techniques excel in scenarios with imbalanced data and contribute to mitigating overfitting \cite{sagi2018ensemble}, primarily due to the disparate feature learning across various models. Given the scarcity of EOPD data, ensemble learning emerges as a promising approach. It improves the predictive accuracy for the young group without degrading the model performance for the elderly group. We independently train five identical models with different initial weights and employ them on the testing set. The median results from these models are selected as the final prediction results for evaluation. 

\begin{figure}[!t]
    \includegraphics[width=\columnwidth]{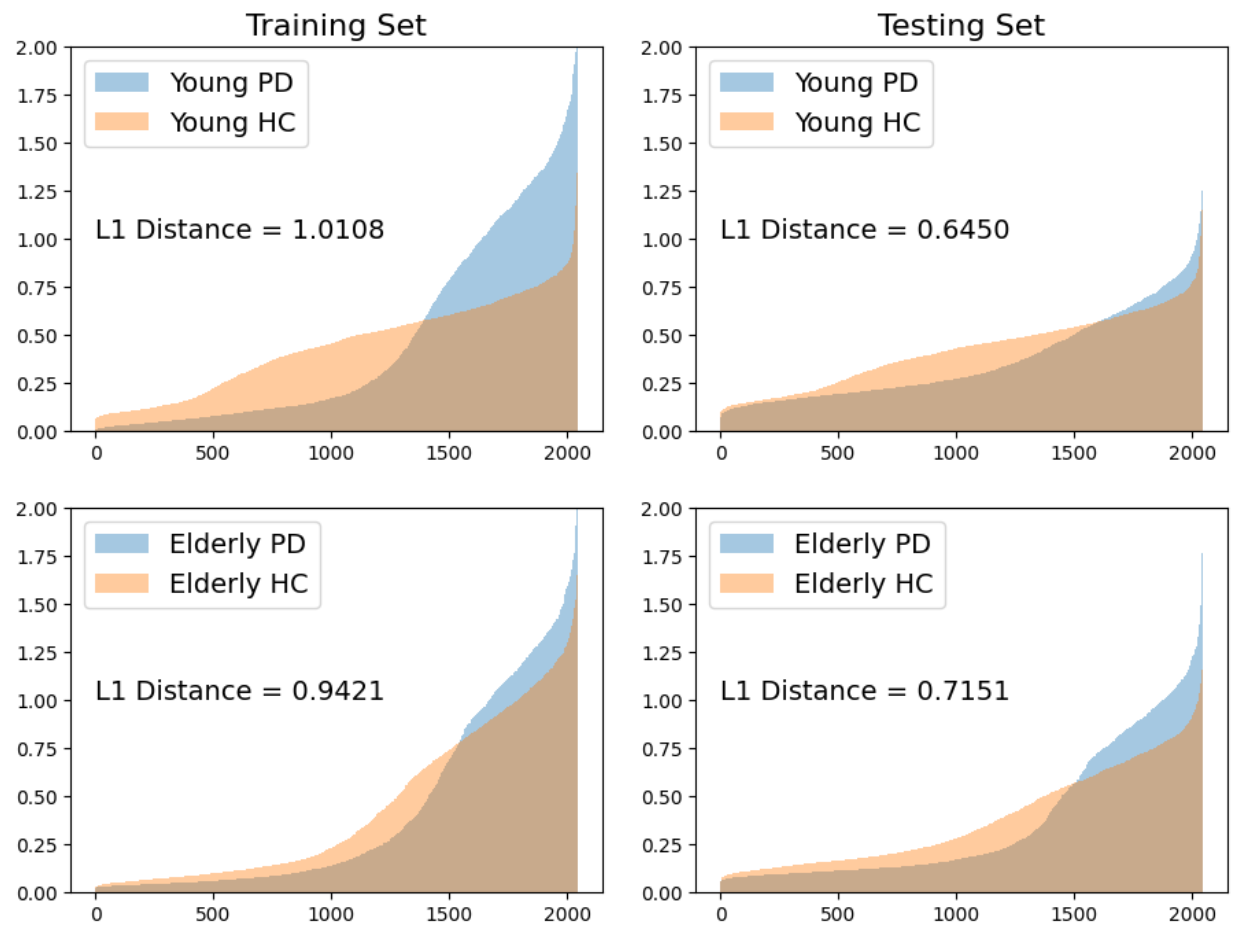}
    \vspace{-15pt}
    \caption{Averaged and Sorted ResNet50 features (with GradCAM-based masking in the training stage): Training Set vs. Testing Set. Compared to Figure \ref{fig1}, the L1 distance in the young group from the training set to the testing set has experienced a smaller decline, indicating that the issue of overfitting has been alleviated.}
    \label{fig4}
\end{figure}

\begin{figure*}[!t]
    \centering
    \includegraphics[width=0.96\textwidth]{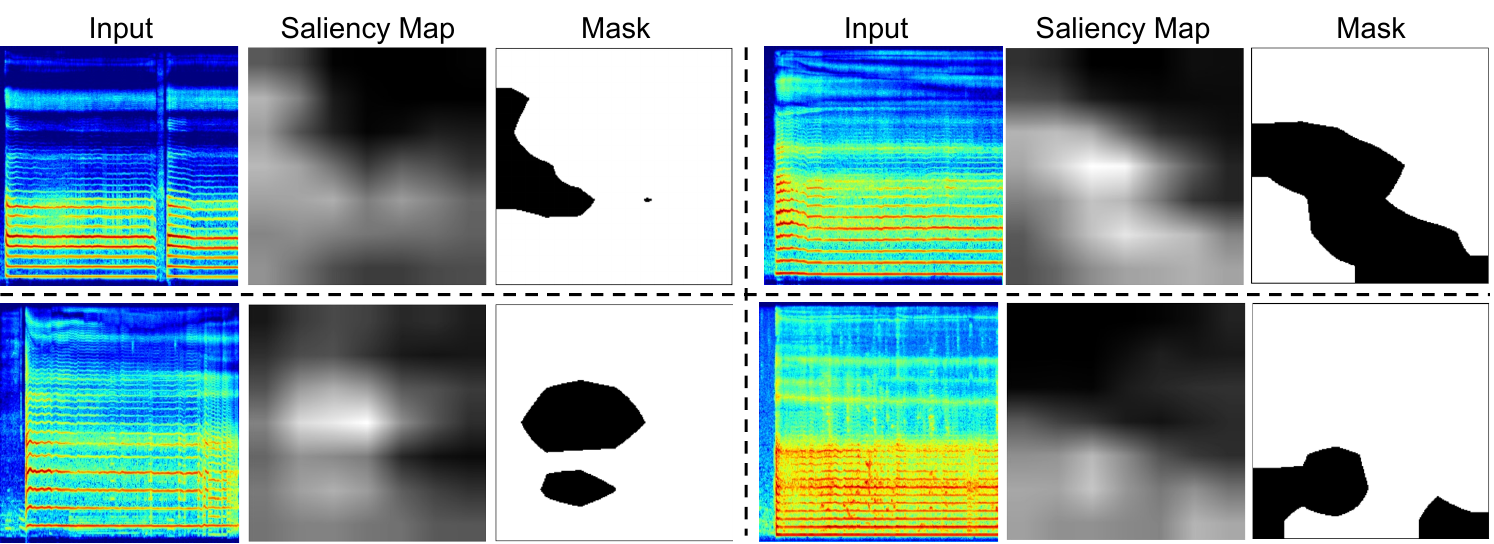}
    \caption{Visualized GradCAM-based masks. \textbf{Left}: Input mel-spectrograms. \textbf{Middle}: Saliency maps. \textbf{Right}: Generated GradCAM-based masks from a ResNet50 model. The black areas highlighted by these masks pinpoint where the model identifies features in the PD data that may be influenced by age. Consequently, applying the learned mask effectively removes this age-related information from the voice input data, resulting in fair predictions for both the young and elderly groups.}
    \label{fig5}
\end{figure*}

\section{Experiments}

In this section, we detail our experimental framework that assesses the efficacy of our proposed GradCAM-based feature masking pipeline.  We investigate the debiasing capabilities of our approach through performance metrics. The mask visualizations further illustrate how the masking targets age-related features in voice data.


\subsection{Experiment Settings}

Our experimental framework revolves around two primary architectures: ResNet50 \cite{he2016deep} and DenseNet161 \cite{huang2017densely}. Following the settings of a previous study \cite{karaman2021robust}, both models have been initialized using ImageNet pre-trained weights. We employed a consistent training regimen for both architectures, spanning 24 epochs. The learning rate was set at $1e^{-5}$, and the batch size was 32. To ensure uniformity across our experiments, all input Mel-spectrograms were resized and reshaped to $3\times224\times224$ dimensions using jet color maps. We set the mask threshold $\theta=0.6$.

In terms of computational infrastructure, ResNet50 models were trained on a Nvidia 2080Ti GPU. The more parameter-intensive DenseNet161 models were trained on a Nvidia 3090 GPU. To evaluate the performance of our models, we adopted the Area Under the Precision-Recall Curve (AUPRC) as our primary evaluation metric. The choice of AUPRC is especially appropriate for a dataset with imbalanced distribution, as it provides a more informative measure\cite{saito2015precision, movahedi2021limitation}.

For our comparative analysis, we employed two debiasing methodologies: the resampling strategy and the adversarial technique \cite{zhang2018mitigating}. In the resampling approach, we deliberately oversampled PD data from the younger groups to ensure that the training data for this group had a PD/HC proportion closely mirroring that of the elderly group. As for the adversarial debiasing technique, we added an additional adversarial age classification module after our CNN architecture. This module introduced an adversarial loss with a weight of $\alpha=0.01$ contingent on the CNN's output when predicting PD/HC. We also set the original biased CNN models and the separate GradCAM-based masking and ensemble models as baselines.

\subsection{Main Results and Analysis.}
The experimental results presented in Table \ref{table3} reveal critical insights into the efficacy of different debiasing strategies. 

When observing the average AUPRC across all models, we find that both ResNet50 and DenseNet161 not only maintain but even slightly enhance their performance when utilizing our proposed method. These results show the capability of our combined GradCAM-based masking and ensemble techniques to improve model fairness without sacrificing accuracy. 

A primary concern in our analysis is the disparity in performance between the young and elderly groups. Compared to the original models, our method dramatically narrows this disparity, registering the most pronounced improvement in fairness across age groups among all baseline methods. Figure \ref{fig4} illustrates the mitigation of the overfitting issue through the application of GradCAM-based masking, wherein the reduction in the level of L1 distance of the young group feature is notably less compared to that observed in Figure \ref{fig1}.

The resampling method, however, fell short in increasing the young group's AUPRC, which indicates its inability to counteract the model's inherent overfitting. This is likely due to the fact that the resampling relies on repeated EOPD data, causing the model to recognize and over-rely on these redundant patterns. Hence, it may fail to generalize well to unseen or new EOPD data.

While the adversarial technique succeeds in reducing disparity, it does so at the expense of the performance within the elderly group, which is not acceptable. It is crucial to maintain high accuracy in the major PD age group to ensure the model’s efficacy and reliability in the clinical application.

In conclusion, the results demonstrate that the GradCAM-based masking technique, especially when paired with ensemble strategies, appears to be a more robust solution for debiasing and ensuring fairness in PD voice detection across different age groups.

\subsection{How does the GradCAM-based Feature Masking Work? A Visualization of the Learned Mask.}
To investigate why and how GradCAM-based feature masking mitigates bias, we present a series of visualizations of GradCAM-based masks extracted from the testing set in Figure \ref{fig5}.

Within the figure, it is apparent that the mask concentrates on some specific frequency regions at different times. These masked areas signify the regions where the model identifies the concentration of features that are influenced by age within the PD data points. These masked regions remove some age-influenced variations in the PD features, playing a crucial role in the model’s debiasing process to overcome the overfitting problem.

\begin{figure}[!t]
    \centering
    \includegraphics[width=0.7\columnwidth]{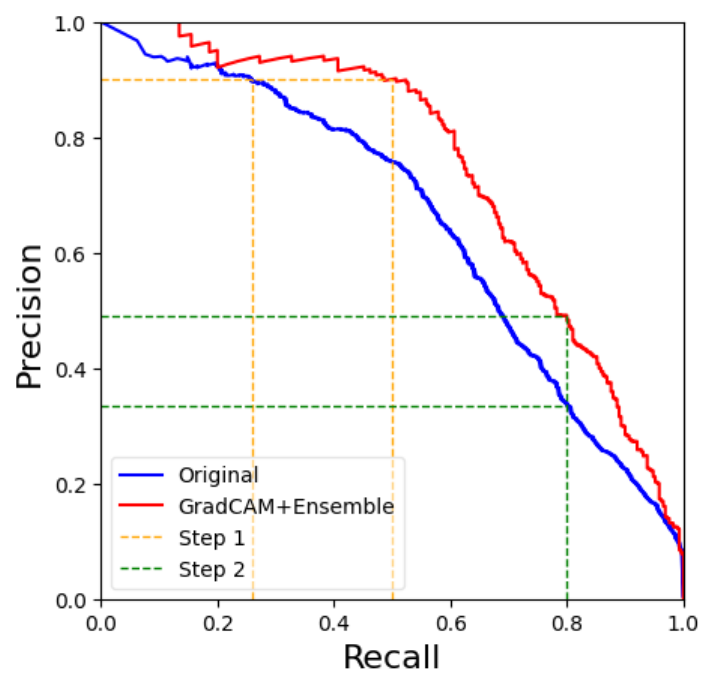}
    \vspace{-12pt}
    \caption{Two-step PD detection strategy. The blue and red curves are the precision-recall curves of the original ResNet50 results and the model trained with the GradCAM-based masking and ensemble method. The \textbf{\color{Goldenrod}{yellow}} and  \textbf{\color{ForestGreen}{green}} dashed lines show the decision points of Step 1 and Step 2, respectively.}
    \label{fig6}
\end{figure}

\section{Two-Step Strategy: Further Improve PD Detection for the Young Group}
Though our proposed methods have enhanced the model's performance on EOPD, there persists a noticeable gap compared to the elderly group, largely attributed to the scarcity of EOPD data. Consequently, for practical clinical applications, we propose a two-step PD detection strategy specifically for the young group, as shown in Figure \ref{fig6}.

In the first stage of this strategy, a model with a large decision value (indicated by the yellow dashed line) is employed to screen young individuals. This initial phase is characterized by high precision, implying a lower rate of false positives, and thereby, those young individuals who receive a positive prediction at this stage are highly probable to be actual PD cases. However, the model at this stage exhibits a comparatively low recall value, a situation which may result in several EOPD cases being missed.

To address the potential omissions in the first stage, the second step employs the same model but with a significantly lower decision value (indicated by the green dashed line) on the previously negative young individuals. This adjustment ensures that the model yields predictions with considerably higher recall, reducing the chances of false negatives. Individuals detected as positive in this phase can be assigned a high-risk indicator, prompting a recommendation for further diverse screening methods for PD.

This strategy significantly amplifies the detection rate of EOPD in clinical scenarios, optimizing both the recall and precision of the model in its respective stages. In addition, as shown in Figure \ref{fig6}, by integrating our proposed GradCAM-Ensemble method, we can realize enhanced recall and precision in each step of the strategy, which is especially crucial for early-onset cases where symptoms may not be as pronounced.

\section{Conclusion and Future Work}
In this paper, we highlight the performance disparity in deep learning models applied to PD voice detection for EOPD and LOPD. We identify the contributing factors to these biases, and in response, we propose a novel debiasing method that combines GradCAM-based feature masking and ensemble modeling to remove age-related features and optimize model capabilities. This innovative approach mitigates biases without compromising the detection performance of either group, demonstrating an effective way to balance sensitivity and precision in PD voice detection. Our two-step detection strategy offers an practical approach to detecting PD in younger populations, providing a more accurate and comprehensive risk assessment that is crucial for early intervention. 

Future research can focus on refining the debiasing approach to ensure better adaptability and performance across a wide range of age groups. Additionally, concurrently addressing various forms of biases, including those related to gender and ethnicity remains a challenge for further exploration.

\bibliographystyle{siam}
\bibliography{ref}
\end{document}